\pdfoutput=1

\documentclass[11pt]{article}

\usepackage[]{emnlp2021}

\usepackage{times}
\usepackage{latexsym}
\usepackage{graphicx}

\usepackage{bm}
\usepackage{amsmath}
\usepackage{amssymb}
\usepackage{amsfonts}

\usepackage{algorithm}
\usepackage{algpseudocode}

\usepackage{array, booktabs}
\usepackage{multirow}
\usepackage{MnSymbol}

\usepackage{threeparttable}

\usepackage[T1]{fontenc}

\usepackage[utf8]{inputenc}

\usepackage{microtype}

\newcommand{\argmax}{\mathop{\rm arg~max}\limits}

\title{Neuro-Symbolic Reinforcement Learning with First-Order Logic}

\author{
Daiki Kimura \:\: Masaki Ono \:\: Subhajit Chaudhury \:\: Ryosuke Kohita \:\: Akifumi Wachi \\
{\bf Don Joven Agravante} \:\: {\bf Michiaki Tatsubori} \:\: {\bf Asim Munawar} \:\: {\bf Alexander Gray} \\
IBM Research \\
\texttt{\small \{daiki, moono, subhajit\}@jp.ibm.com, \{kohi, akifumi.wachi\}@ibm.com} \\[-3pt]
\texttt{\small don.joven.r.agravante@ibm.com, mich@jp.ibm.com, \{asim, alexander.gray\}@ibm.com}
}

\begin{document}
\maketitle
\begin{abstract}
Deep reinforcement learning (RL) methods often require many trials before convergence, and no direct interpretability of trained policies is provided. In order to achieve fast convergence and interpretability for the policy in RL, we propose a novel RL method for text-based games with a recent neuro-symbolic framework called Logical Neural Network, which can learn symbolic and interpretable rules in their differentiable network. The method is first to extract first-order logical facts from text observation and external word meaning network (ConceptNet), then train a policy in the network with directly interpretable logical operators. Our experimental results show RL training with the proposed method converges significantly faster than other state-of-the-art neuro-symbolic methods in a TextWorld benchmark.
\end{abstract}

\section{Introduction}

Deep reinforcement learning~(RL) has been successfully applied to many applications, such as computer games, text-based games, and robot control applications~\cite{dqn,Narasimhan+:EMNLP2015:LSTM-DQN,kimura2018daqn,Yuan+:2018:LSTM-DRQN,marioirl,kimura2021reinforcement}. 
However, these methods require many training trials for converging to the optimal action policy, and the trained action policy is not understandable for human operators. This is because, although the training results are sufficient, the policy is stored in a black-box deep neural network. These issues become critical problems when the human operator wants to solve a real-world problem and verify the trained rules. If the trained rules are understandable and modifiable, the human operator can control them and design an action restriction. While using a symbolic~(logical) format as representation for stored rules is suitable for achieving interpretability and quick training, it is difficult to train the logical rules with a traditional training approach.

\begin{figure}[tb]
\centering
\includegraphics[width=1.0\linewidth]{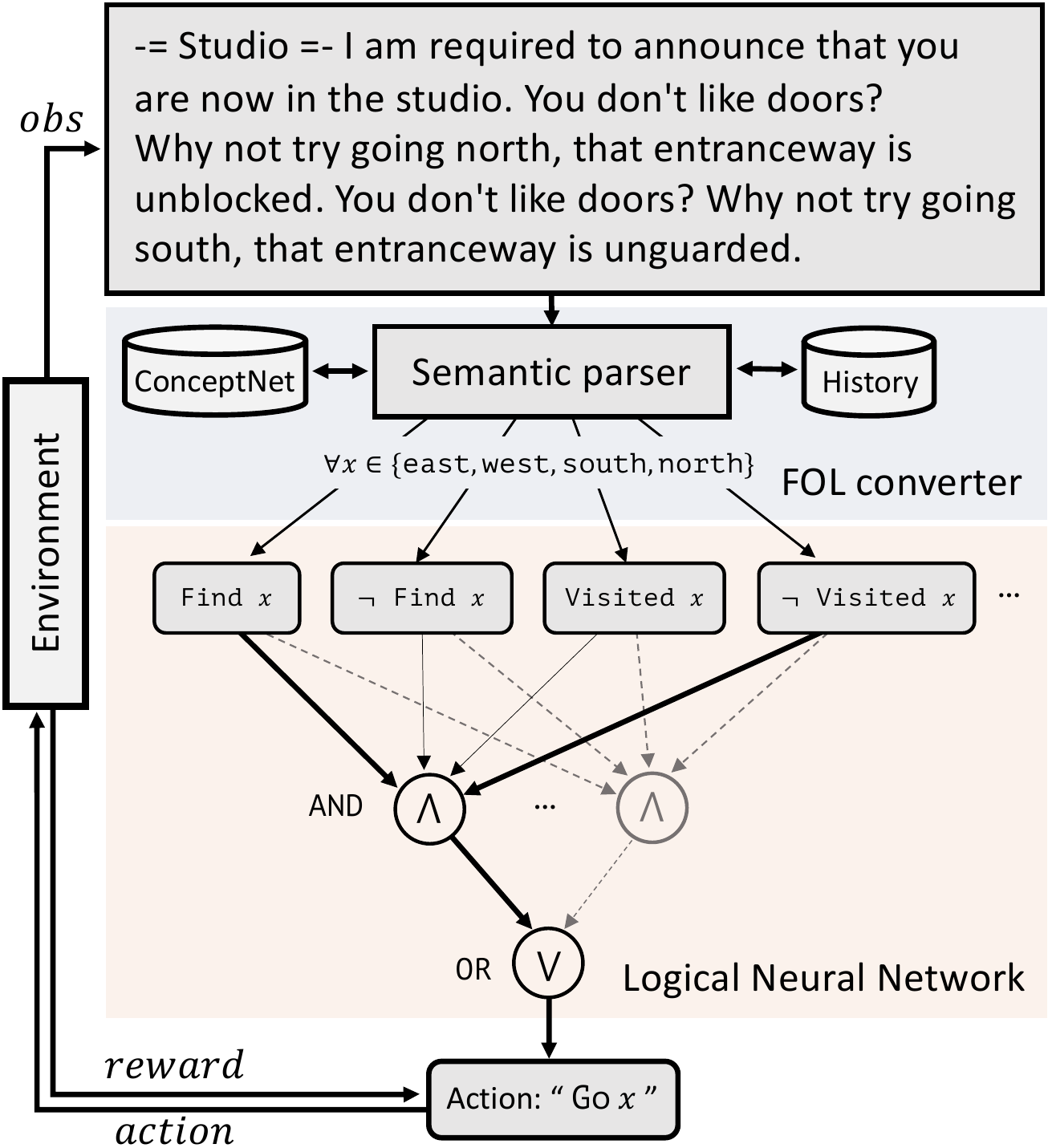}
\caption{Overview of the proposed method. The agent takes a text observation from the environment, and the first-order logical facts are extracted from an FOL converter that uses a semantic parser, ConceptNet, and history. The weights~(shown by line thickness in this figure) of the network are updated by these extracted predicate logics. Solid lines show one trained rule; when the agent finds a direction~$x$ and the direction~$x$ has not been visited, the agent takes a ``Go~$x$'' action. Dashed lines show the initial connections before training.}
\label{fig:overview}
\end{figure}

In order to train logical rules, a recent neuro-symbolic framework called the Logical Neural Network~(LNN)~\cite{riegel2020logical} has been proposed to simultaneously provide key properties of both the neural network~(learning) and the symbolic logic~(reasoning). The LNN can train the symbolic rules with logical functions in the neural networks by having an end-to-end differentiable network minimizes a contradiction loss. Every neuron in the LNN has a component for a formula of weighted real-valued logics from a unique logical conjunction, disjunction, or negation nodes, and then it can calculate the probability and logical contradiction loss during the inference and training. At the same time, the trained LNN can extract obtained logical rules by selecting high weighted connections that represent the important rules for an action policy.

In this paper, we propose an action knowledge acquisition method featuring a neuro-symbolic LNN framework for the RL algorithm. Through experiments, we demonstrate the advantages of the proposed method for real-world problems which is not logically grounded games such as Blocks World. Since natural language observation is easier to convert into logical information than visual or audio, we tackle text-based interaction games for verifying the proposed method.

Figure~\ref{fig:overview} shows an overview of our method. The observation text is input to a semantic parser to extract the logical values of each propositional logic. In this case, the semantic parser finds there are two exits~(north and south). The method then converts first-order logical~(predicates) facts from the propositional logics and categories of each word, such as $\exists x \in \{\text{south}, \text{north}\}, \langle \text{find } x \rangle=True$ and $\exists x \in \{\text{east}, \text{west}\}, \langle \text{find } x \rangle=False$. These extracted predicated logics are fed into LNN which has some conjunction gates and one disjunction gate. The LNN trains the weights for these connections by the reward value to obtain the action policy. 

The contributions of this paper are as follows. 
\begin{itemize}
\item The paper describes design and implementation of a novel neuro-symbolic RL for a text-based interaction games.
\item The paper explains an algorithm to extract first-order logical facts from given textual observation by using the agent history and ConceptNet as an external knowledge.
\item We observed our proposed method has advantages for faster convergence and interpretability than state-of-the-art methods and baselines by ablation study on the text-based games.  
\end{itemize}

\section{Related work}

Various prior works have examined RL for text-based games. LSTM-DQN~\cite{Narasimhan+:EMNLP2015:LSTM-DQN} is an early study on an LSTM-based encoder for feature extraction from observation and Q-learning for action policy. LSTM-DQN++~\cite{Yuan+:2018:LSTM-DRQN} extended the exploration and LSTM-DRQN~\cite{Yuan+:2018:LSTM-DRQN} was proposed for adding memory units in the action scorer. KG-DQN~\cite{Ammanabrolu+Riedl:NAACL2019:KG-DQN} and GATA~\cite{Adhikari+:NeurIPS2020:GATA} extended the language understanding. LeDeepChef~\cite{adolphs2020ledeepchef} used recurrent feature extraction along with the A2C~\cite{mnih2016asynchronous}. CREST~\cite{Chaudhury+:EMNLP2020:CREST} was proposed for pruning observation information. TWC~\cite{Murugesan+:AAAI2021:TWC} was proposed for utilizing common sense reasoning. However, none of these studies used the neuro-symbolic approach. 

For recent neuro-symbolic RL work, the Neural Logic Machine~(NLM)~\cite{dong2018neural} was proposed as a method for combination of deep neural network and symbolic logic reasoning. It uses a sequence of multi-layer perceptron layers to deduct symbolic logics. Rules are combined or separated during forward propagation, and an output of the entire architecture represents complicated rules. In this paper, we compare our method with this NLM.

\section{Proposed method}

\subsection{Problem formulation}

As text-based games are sequential decision-making problems, they can naturally be applied to RL. These games are partially observable Markov decision processes~(POMDP)~\cite{KAELBLING199899}, where the observation text does not include the entire information of the environment. Formally, the game is a discrete-time POMDP defined by~$\langle S,A,T,R,\omega,O,\gamma \rangle$, where $S$~is a set of states~($s_t \in S$), $A$~is a set of actions, $T$~is a set of transition probabilities, $R$~is a reward function, $\omega$~is a set of observations~($o_t \in \omega$), $O$~is a set of conditional observation probabilities, and $\gamma$~is a discount factor. Although the state~$s_t$ contains the complete internal information, the observation~$o_t$ does not. In this paper, we follow following two assumptions: one, the word in each command is taken from a fixed vocabulary~$V$, and two, each action command consists of two words~(verb and object). The objective for the agent is to maximize the expected discounted reward~$E[\sum_t \gamma^t r_t]$.

\subsection{Method}

The proposed method consists of two processes: converting text into first-order logic~(FOL), and training the action policy in LNN. 

\subsubsection{FOL converter}

The FOL converter converts a given natural observation text~$o_t$ and observation history~$(o_{t-1}, o_{t-2}, ...)$ into first-order logic facts. The method first converts text into propositional logics~$l_{i,t}$ by a semantic parser from $o_t$, such as, the agent understands an opened direction from the current room. The agent then retrieves the class type~$c$ of the word meaning in propositional logic~$l_{i,t}$ by using ConceptNet~\cite{10.1023/B:BTTJ.0000047600.45421.6d} or the network of another word's definition. For example, ``east'' and ``west'' are classified as a {\it direction}-type, and ``coin'' is as a {\it money}-type. The class is used for selecting the appropriate LNN for FOL training and inference.

\subsubsection{LNN training}

The LNN training component is for obtaining an action policy from the given FOL logics. LNN~\cite{riegel2020logical} has logical conjunction~(AND), logical disjunction~(OR), and negation~(NOT) nodes directly in its neural network. In our method, we prepare an AND-OR network for training arbitrary rules from given inputs. As shown in Fig.~\ref{fig:overview}, we prepare all logical facts at the first layer, several AND gates (as many as the network is required) at the second layer, and one OR gate connected to all previous AND gates. During the training, the reward value is used for adding a new AND gate, and for updating the weight value for each connection. More specifically, the method is storing the replay buffer which has current observation~$o_t$, action~$a_t$, reward~$r_t$, and next observation~$o_{t+1}$ value. For each training step, the method selects some replies, and it extracts first-order logical facts from current observation~$o_t$ and action~$a_t$. The LNN trains by this fact inputs and reward; that means it forwards from input facts through LNN, calculates a loss values from the reward value, and optimizes weights in LNN. The whole training mechanism is similar to DQN~\cite{mnih2013playing}, the difference from these is the network. To aid the interpretability of node values, we define a threshold~$\alpha \in [\frac{1}{2},1]$ such that a continuous value is considered {\it True} if the value is in $[\alpha, 1]$, and {\it False} if it is in $[0, 1-\alpha]$. 

Algorithm~\ref{alg:proposed_method} describes the whole algorithm for the proposed method.

\begin{algorithm}[tb]
\caption{RL by FOL-LNN}\label{euclid} 
\label{alg:proposed_method}
\begin{algorithmic}[1]
\Procedure{Reinforcement Learning}{}
\For{$t = 1, 2, 3, ...$}
\State $ o_t \leftarrow \text{Observe } \textit{observation}$
\State $ l_{t,i} \leftarrow \text{Extract } \textit{logic} \text{ from} \: o_t, o_{t-1}, ...$
\For{$i = 1, 2, 3, ...$}
\State $ c \leftarrow \text{Find } \textit{class} \text{ from ConceptNet}$
\State $ \theta^c \leftarrow \text{Select } \textit{LNN}$
\State $ l_{t,i}^c \leftarrow \text{Convert into } \textit{FOL logic}$
\State $ a_{t,i} \leftarrow \theta^c(l_{t,i}^c)$
\EndFor{}
\State $ a_{t} \leftarrow \argmax a_{t,i}$
\State $ r_{t}, o_{t+1} \leftarrow \text{Get } \textit{reward} \text{ and } \textit{next obs}$
\State $ \text{Store reply } \langle o_t, a_t, r_{t},o_{t+1} \rangle$
\State $ \nabla\theta \leftarrow \text{Update LNN from reply}$
\EndFor{}
\EndProcedure
\end{algorithmic}
\end{algorithm}

\tabcolsep = 0.9mm
\renewcommand\arraystretch{1.4}
{
\begin{table*}[t]
\caption{Average reward and number of steps~({\bf reward}: higher is better / {\bf number of steps}: lower is better) for each epoch on 50~unseen games with three difficulty levels. These results are from moving average~($N=100$) and 5~random seeds. Training is done on only small-size games. Although neuro-only method cannot solve unseen test games, our proposed method~(FOL-LNN) can solve and converge extremely faster than other SOTAs and baselines.}
\begin{center}
{\small
\begin{threeparttable}[h]

\begin{tabular}{c|rrr||rrr||rrr}
\toprule
& \multicolumn{3}{c||}{Easy game} & \multicolumn{3}{c||}{Medium game} & \multicolumn{3}{c}{Hard game}\\

Epoch & \multicolumn{1}{c}{100} & \multicolumn{1}{c}{200} & \multicolumn{1}{c||}{1000} & \multicolumn{1}{c}{100} & \multicolumn{1}{c}{200} & \multicolumn{1}{c||}{2000} & \multicolumn{1}{c}{100} & \multicolumn{1}{c}{200} & \multicolumn{1}{c}{2000}\\
\midrule
\midrule
LSTM-DQN++~\tnote{*} & 0.07\hspace{1.5pt}/\hspace{1.5pt}93.4 & 0.10\hspace{1.5pt}/\hspace{1.5pt}90.9 & 0.12\hspace{1.5pt}/\hspace{1.5pt}88.6 & 0.00\hspace{1.5pt}/\hspace{1.5pt}99.9 & 0.00\hspace{1.5pt}/\hspace{1.5pt}99.6 & 0.03\hspace{1.5pt}/\hspace{1.5pt}97.3 & 0.00\hspace{1.5pt}/\hspace{1.5pt}99.9 & 0.00\hspace{1.5pt}/\hspace{1.5pt}99.9 & 0.04\hspace{1.5pt}/\hspace{1.5pt}96.6\\
\midrule
NLM-DQN~\tnote{**} & 0.87\hspace{1.5pt}/\hspace{1.5pt}26.4 & 0.93\hspace{1.5pt}/\hspace{1.5pt}20.8 & {\bf 1.00}\hspace{1.5pt}/\hspace{1.5pt}{\bf 15.0} & 0.27\hspace{1.5pt}/\hspace{1.5pt}81.1 & 0.48\hspace{1.5pt}/\hspace{1.5pt}65.9 & {\bf 1.00}\hspace{1.5pt}/\hspace{1.5pt}29.7 & 0.01\hspace{1.5pt}/\hspace{1.5pt}99.7 & 0.10\hspace{1.5pt}/\hspace{1.5pt}94.8 & 0.66\hspace{1.5pt}/\hspace{1.5pt}64.0\\
\hline
NN-DQN & 0.91\hspace{1.5pt}/\hspace{1.5pt}22.8 & 0.95\hspace{1.5pt}/\hspace{1.5pt}19.0 & {\bf 1.00}\hspace{1.5pt}/\hspace{1.5pt}{\bf 15.0} & 0.48\hspace{1.5pt}/\hspace{1.5pt}65.6 & 0.65\hspace{1.5pt}/\hspace{1.5pt}54.3 & {\bf 1.00}\hspace{1.5pt}/\hspace{1.5pt}29.5 & 0.19\hspace{1.5pt}/\hspace{1.5pt}89.0 & 0.28\hspace{1.5pt}/\hspace{1.5pt}84.0 & 0.97\hspace{1.5pt}/\hspace{1.5pt}46.3\\
\hline
LNN-NN-DQN & 0.88\hspace{1.5pt}/\hspace{1.5pt}24.8 & 0.94\hspace{1.5pt}/\hspace{1.5pt}20.2 & {\bf 1.00}\hspace{1.5pt}/\hspace{1.5pt}{\bf 15.0} & 0.49\hspace{1.5pt}/\hspace{1.5pt}65.8 & 0.61\hspace{1.5pt}/\hspace{1.5pt}57.0 & {\bf 1.00}\hspace{1.5pt}/\hspace{1.5pt}29.6 & 0.24\hspace{1.5pt}/\hspace{1.5pt}86.9 & 0.27\hspace{1.5pt}/\hspace{1.5pt}84.9 & 0.97\hspace{1.5pt}/\hspace{1.5pt}47.4\\
\hline
FOL-LNN~{\bf(Ours)} & {\bf 0.95}\hspace{1.5pt}/\hspace{1.5pt}{\bf 19.0} & {\bf 0.98}\hspace{1.5pt}/\hspace{1.5pt}{\bf 17.1} & {\bf 1.00}\hspace{1.5pt}/\hspace{1.5pt}{\bf 15.0} & {\bf 0.94}\hspace{1.5pt}/\hspace{1.5pt}{\bf 32.7} & {\bf 0.97}\hspace{1.5pt}/\hspace{1.5pt}{\bf 30.7} & {\bf 1.00}\hspace{1.5pt}/\hspace{1.5pt}{\bf 28.6} & {\bf 0.95}\hspace{1.5pt}/\hspace{1.5pt}{\bf 44.8} & {\bf 0.98}\hspace{1.5pt}/\hspace{1.5pt}{\bf 43.5} & {\bf 1.00}\hspace{1.5pt}/\hspace{1.5pt}{\bf 42.0}\\
\bottomrule

\end{tabular}

\begin{tablenotes}
\item[*] State-of-the-art neuro-only method with a simple DQN action scorer~\cite{Narasimhan+:EMNLP2015:LSTM-DQN}
\item[**] State-of-the-art neuro-symbolic method has same input as ours and other neuro-symbolic methods~\cite{dong2018neural}
\end{tablenotes}
\end{threeparttable}
}
\end{center}
\label{tab:result}
\end{table*}
}

\section{Experiments}

We evaluated the proposed method on a coin-collector game in TextWorld~\cite{Cote+:CGW2018:TextWorld} with three different difficulties~({\it easy}, {\it medium}, and {\it hard}). The objective of the game is to find and collect a coin which is placed in a room within connected rooms. Since we tackle a real-world game problem rather than a symbolic games, we need to extract logical facts from given natural texts for neuro-symbolic methods. We prepare the following propositional logics as extracting logical facts: {\it which object is found in the observation}, {\it which direction has already been visited}, and {\it which direction the agent comes from initially}. These logical values are easily calculated from visited room history and word definitions. In this experiment, we prepared 26~logical values\footnote{$26 = (5$~(object)~$+ 4$~(visited)~$+ 4$~(initial)$)\times 2$~(negation)}, and all the following neuro-symbolic methods used these value as input. For the evaluation metric, we focused on (1)~the test reward value on the unseen~(test) games and (2)~the number of steps to achieve the goal on unseen games. Since we focus on the performance of generalization, we only use 50~small-size~($\text{level}=5$) games for training, 50~unseen games from 5 different size~($\text{level} = 5, 10, 15, 20, 25$) games for test\footnote{The agent needs to generalize the game level size}, and mini-batch in training~(batch size~$= 4$). The other parameters for the game and agent follow LSTM-DQN++~\cite{Narasimhan+:EMNLP2015:LSTM-DQN}.

\newpage
We prepared five methods for an evaluation of the proposed method: 

\begin{itemize}
\item LSTM-DQN++~\cite{Narasimhan+:EMNLP2015:LSTM-DQN}: State-of-the-art neuro-only method with a simple DQN action scorer. We use this method as a baseline method for the neuro-only agent, and LSTM receives extracted embedding vector from natural text information.  
\item NLM-DQN~\cite{dong2018neural}: State-of-the-art neuro-symbolic method. The input is propositional logical values that is also used in following baselines and proposed method. The original NLM uses the REINFORCE~\cite{williams1992simple} algorithm, but in order to handle text-based games with the same setting as the other methods, we applied the DQN algorithm. In short, the method uses an NLM layer instead of an LSTM~\cite{hochreiter1997long} for the encoder of the LSTM-DQN++ method. We tuned the hyper-parameters from the same search space as the original paper. 
\item NN-DQN: Na\"ive neuro-symbolic baseline method. The input of the network is propositional logical values, and it uses a multi layer perceptron as the encoder of the LSTM-DQN++. 
\item LNN-NN-DQN: Neuro-symbolic baseline method. The method first gets propositional logical values, it converts by LNN into some conjunction values for all combinations of given logical values, and then it inputs them into a multi layer perceptron. It differs from NN-DQN in that LNN-NN-DQN has prepared conjunction nodes, which should lead to faster training in beginning of the training, and better interpretabiliity after the training.
\item FOL-LNN: Our neuro-symbolic method.
\end{itemize}

\begin{figure}[t]
\begin{minipage}[t]{0.47\hsize}
\begin{center}
\includegraphics[width=1.0\linewidth]{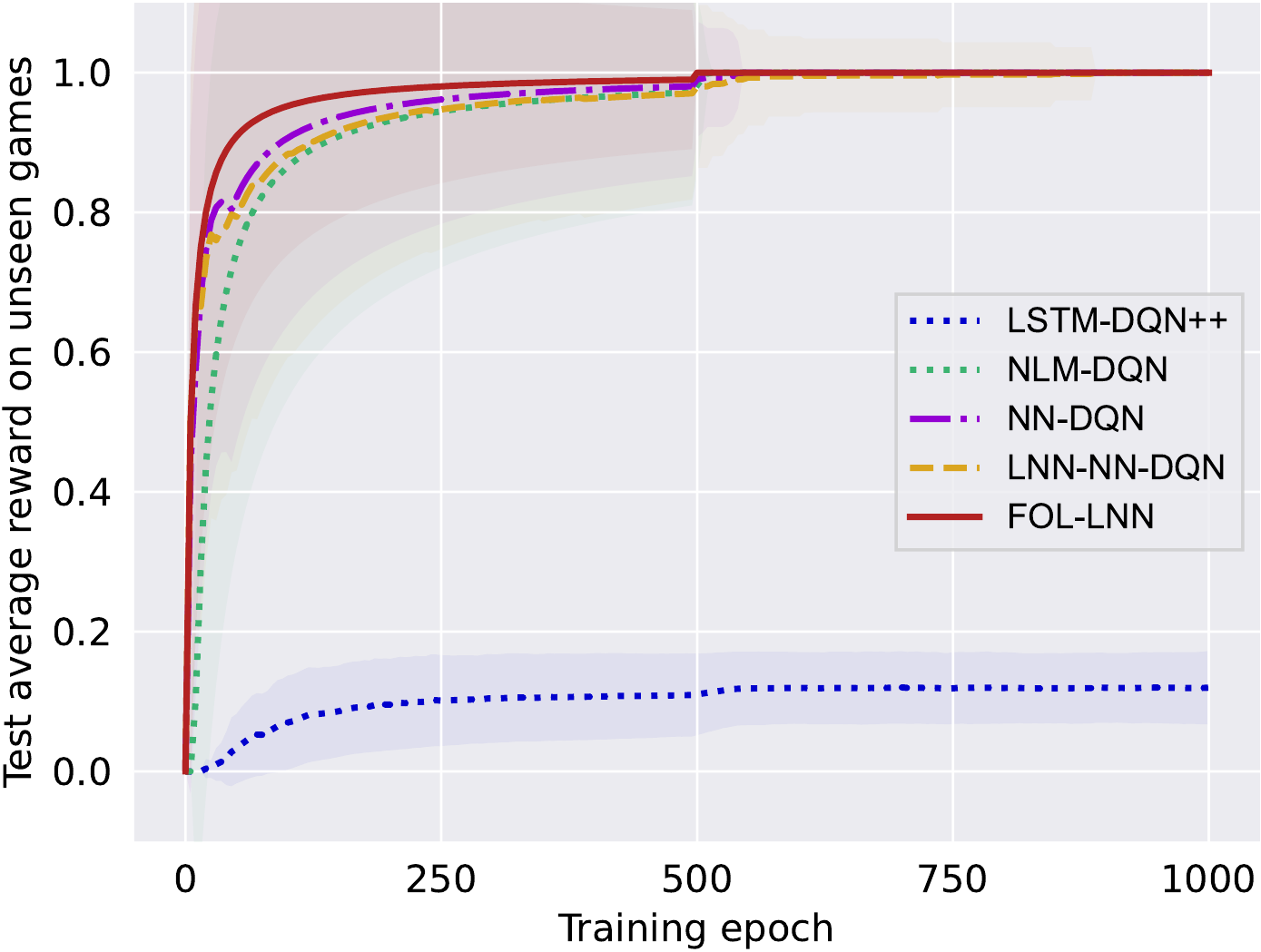}
\caption*{(a) Easy, Test reward}
\end{center}
\end{minipage}
\begin{minipage}[t]{0.47\hsize}
\begin{center}
\includegraphics[width=1.0\linewidth]{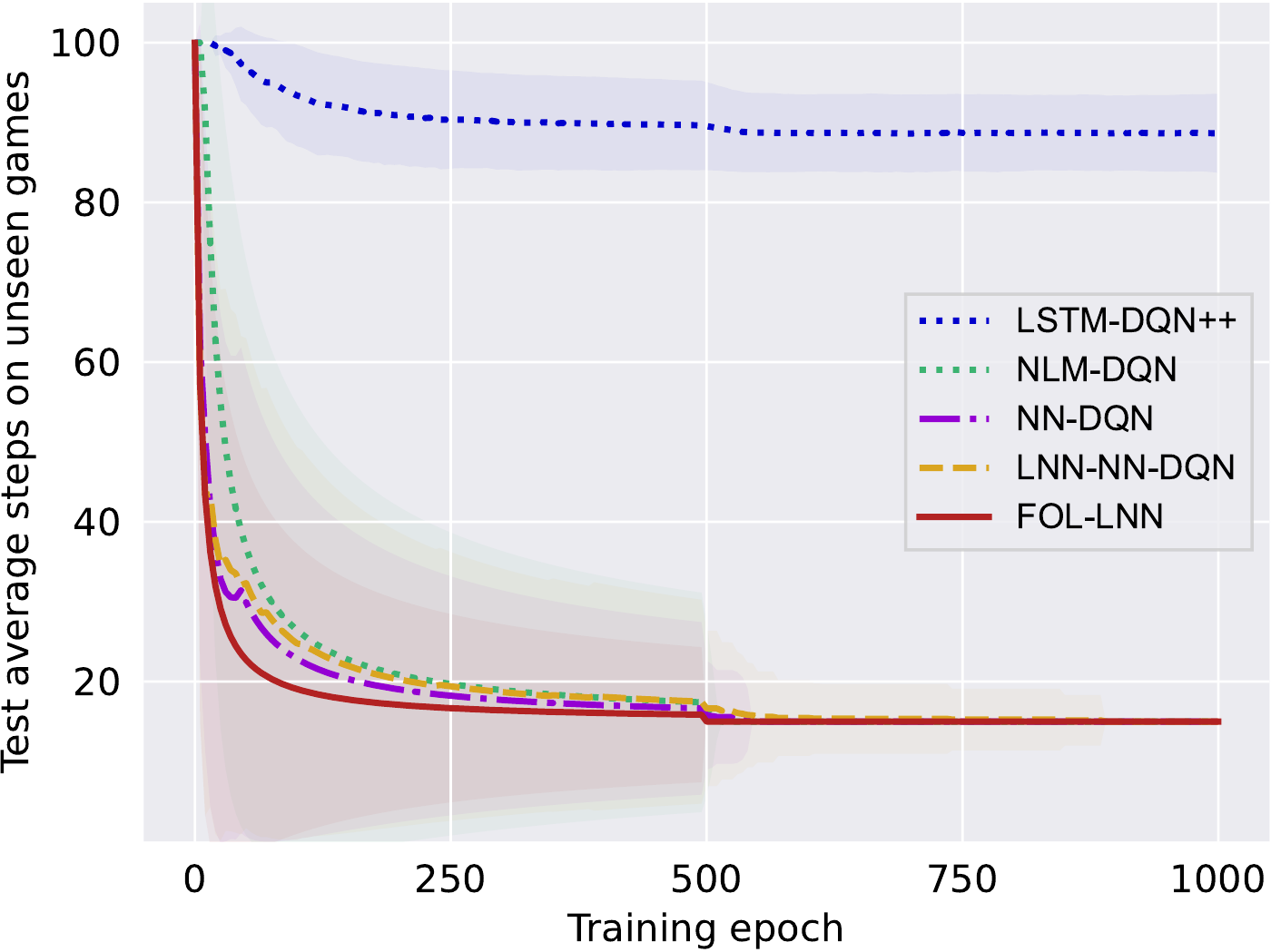}
\caption*{(b) Easy, Test step}
\end{center}
\end{minipage}
\\
\vspace{5pt}
\\
\begin{minipage}[t]{0.47\hsize}
\begin{center}
\includegraphics[width=1.0\linewidth]{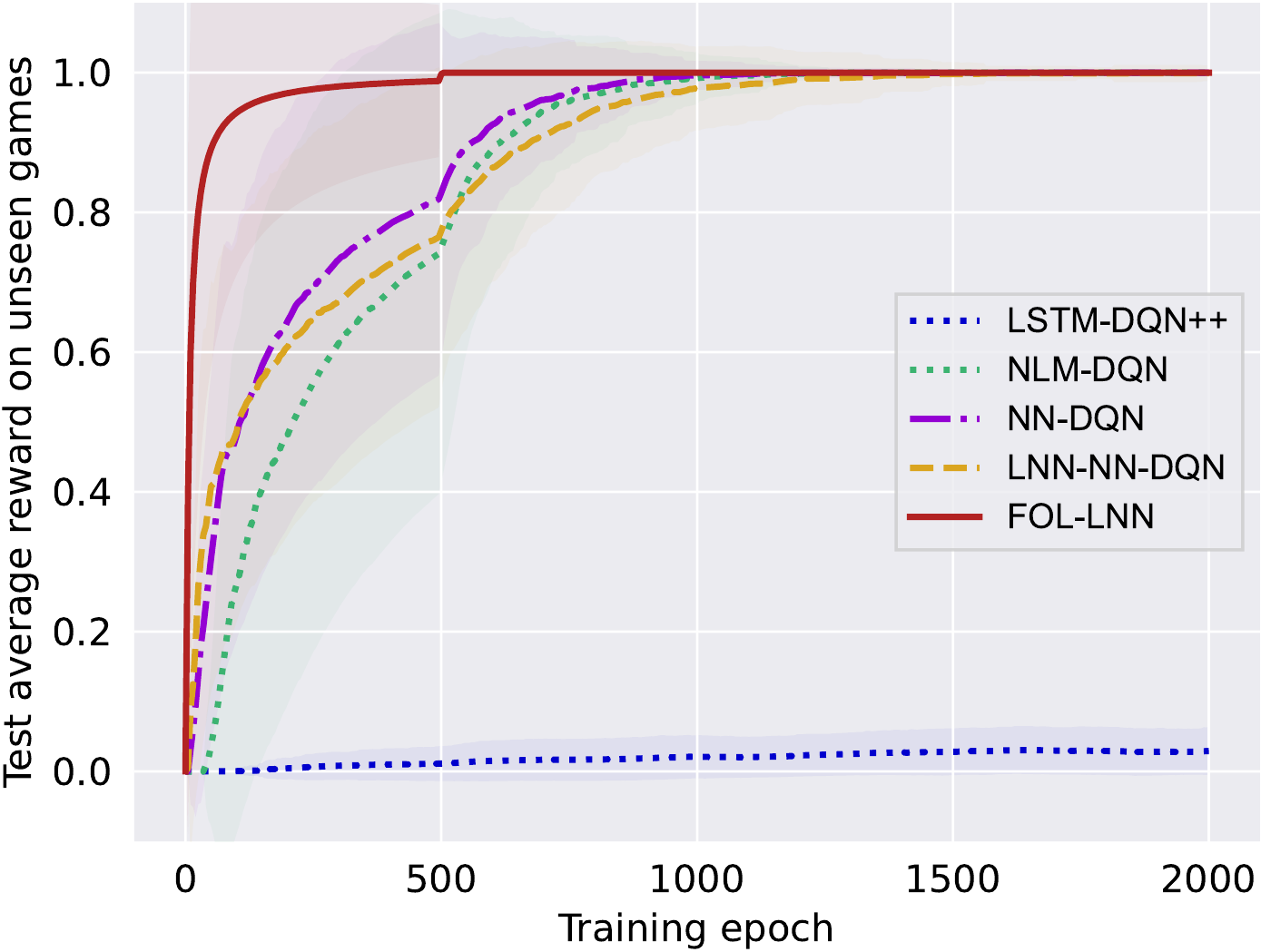}
\caption*{(c) Medium, Test reward}
\end{center}
\end{minipage}
\begin{minipage}[t]{0.47\hsize}
\begin{center}
\includegraphics[width=1.0\linewidth]{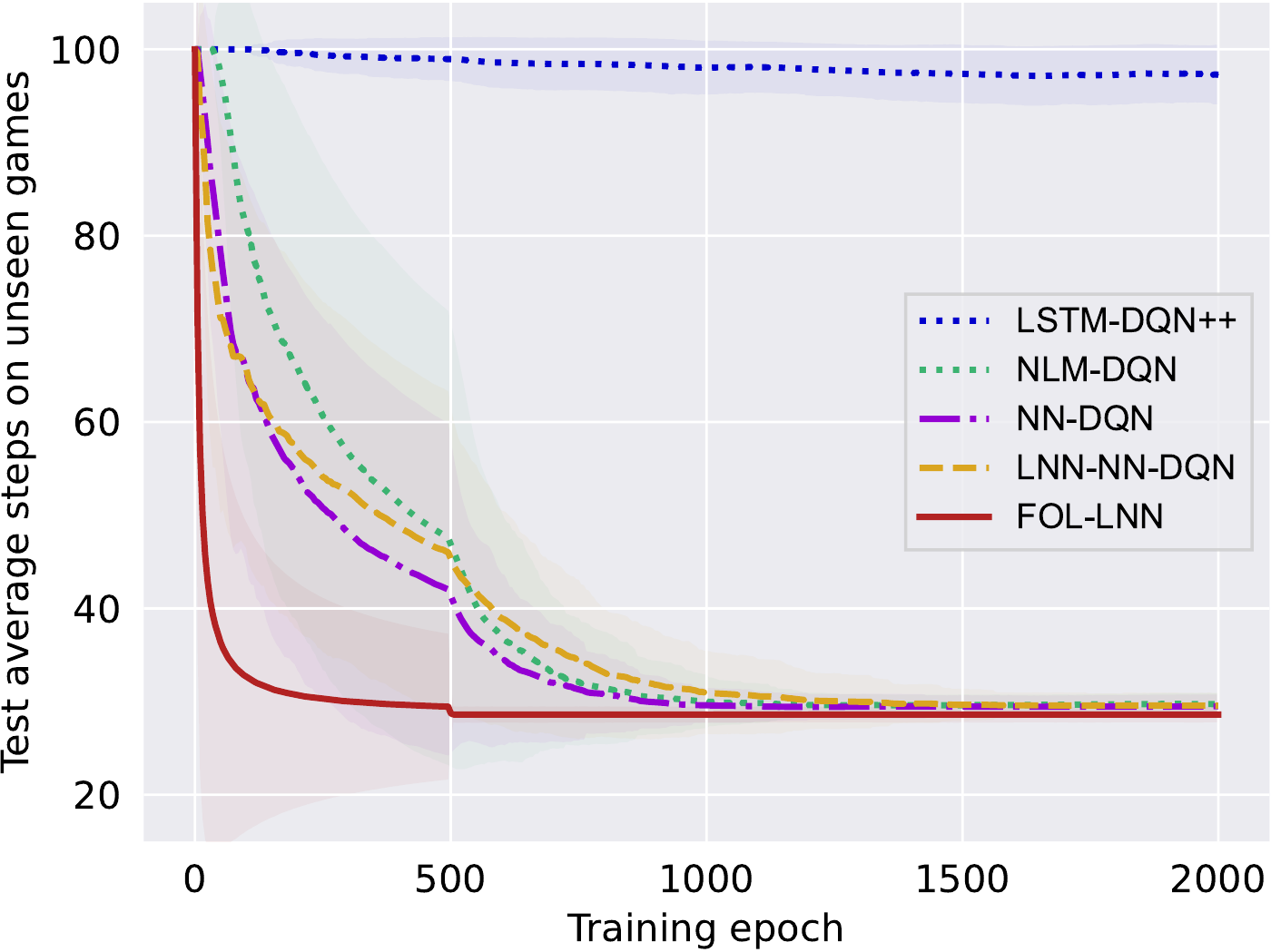}
\caption*{(d) Medium, Test step}
\end{center}
\end{minipage}
\\
\vspace{5pt}
\\
\begin{minipage}[t]{0.47\hsize}
\begin{center}
\includegraphics[width=1.0\linewidth]{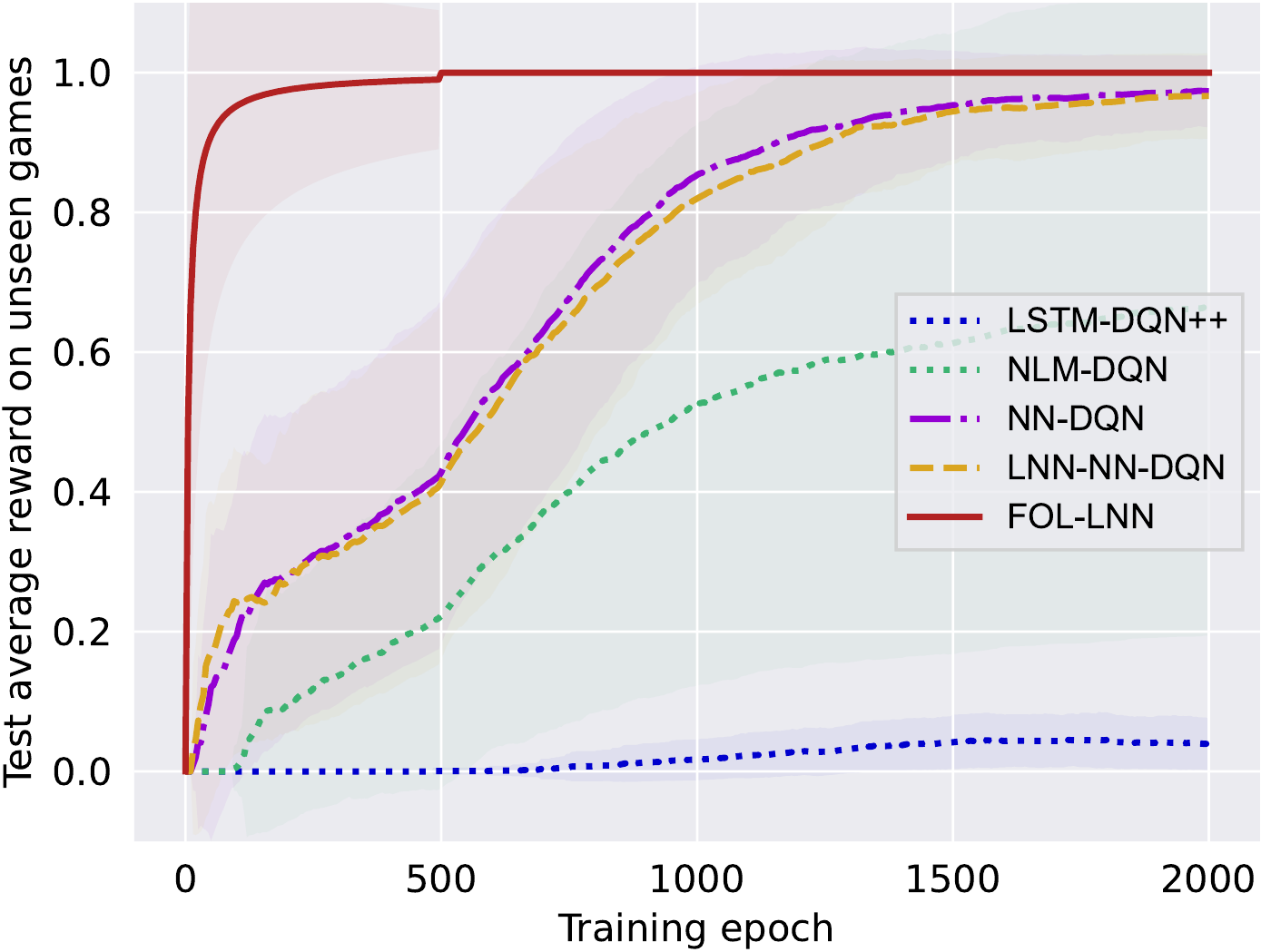}
\caption*{(e) Hard, Test reward}
\end{center}
\end{minipage}
\begin{minipage}[t]{0.47\hsize}
\begin{center}
\includegraphics[width=1.0\linewidth]{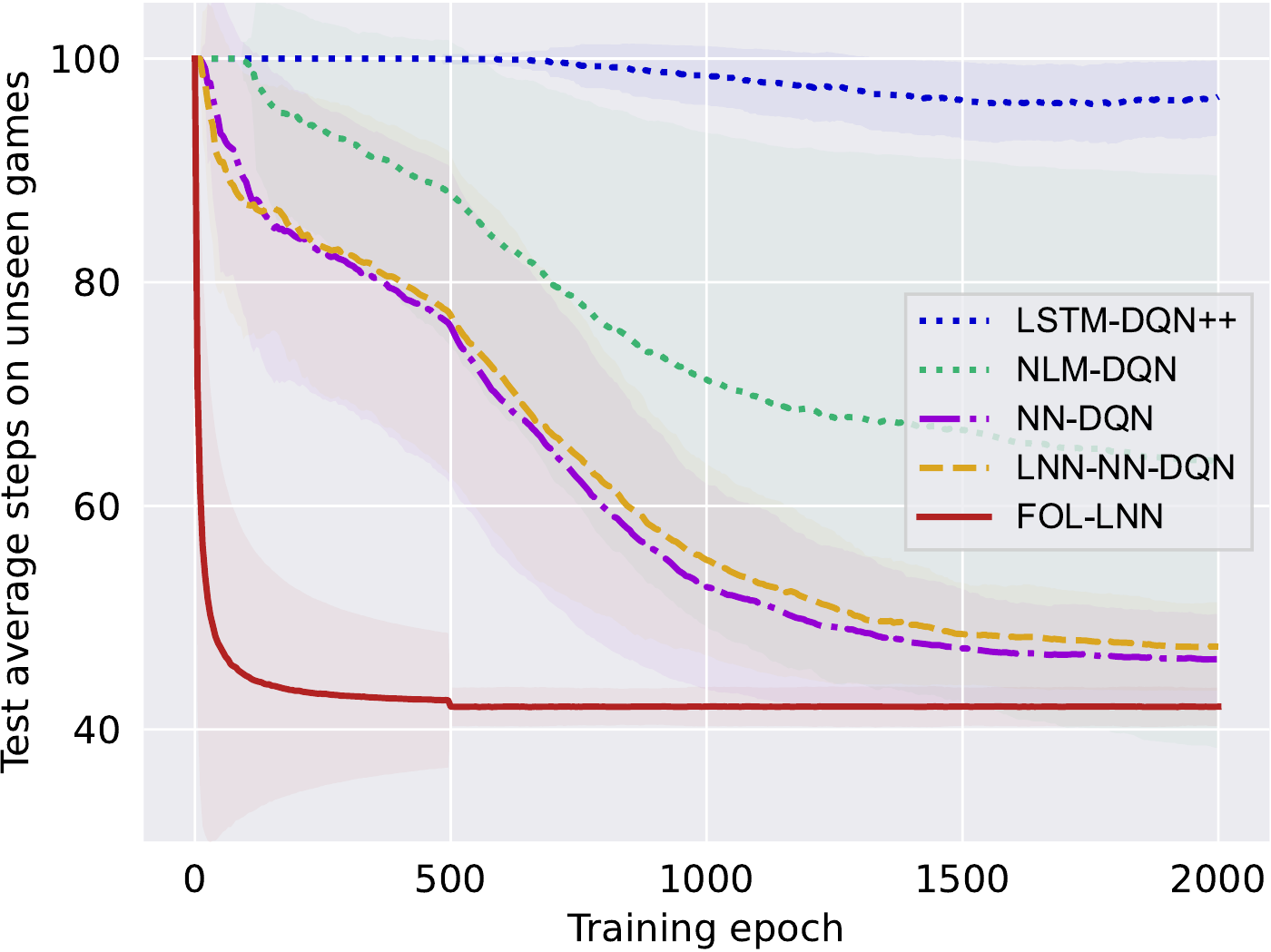}
\caption*{(f) Hard, Test step}
\end{center}
\end{minipage}
\caption{Curves for reward and number of steps for 50~unseen games. Moving average is applied.}
\label{fig:result} 
\end{figure}

Table~\ref{tab:result} shows the test reward and test step values on unseen games, and Fig.~\ref{fig:result} shows curves. First, all the RL results with logical input were better than those with textual input. Second, our proposed method could converge much faster than the other neuro-symbolic state-of-the-art and baseline methods. Third, only our method could extract the trained rules by checking the weight value of the LNN. We attached the extracted rules from the {\it medium} level games here:

{\small
\begin{eqnarray*}
\lefteqn{\exists x \in W_{\text{{\it money}}}} \\
&\langle \text{find } x \rangle \to \llangle \text{take } x \rrangle,
\\[5pt]
\lefteqn{\exists x \in W_{\text{{\it direction}}}} \\
&(\langle \text{find } x \rangle \land \lnot \langle \text{visited } x \rangle \land \lnot \langle \text{initial } x \rangle) \lor & \\
&(\langle \text{find } x \rangle \land \langle \text{all are visited} \rangle \land \langle \text{initial } x \rangle) & \to  \llangle \text{go } x \rrangle , \end{eqnarray*}
}

\noindent where $W_{\text{{\it direction}}}$ is a set of words in a type of ``direction'' in ConceptNet. The rule for "take"-action is for taking a coin. The first conjunction rule for ``go''-action is for visiting an un-visited room, and the second rule is for returning to the initial room from a dead-end. With our proposed method, we can see that these trained rules will be helpful for operating the neural agent in real use cases.

\section{Conclusion}

In this paper, we proposed a novel neuro-symbolic method for RL on text-based games. According to the evaluation on the natural language text-based game with several difficulties, our method can converge extremely faster than other state-of-the-art neuro-only and neuro-symbolic methods, and extract trained logical rules for improving interpretability of the model. 

\section*{Discussion about ethics}

Our model is not using any sensitive contexts such as legal or health-care settings. The data set used in our experiment does not contain any sensitive information. Since our proposed neuro-symbolic RL method can extract the trained rules for interpretability of the model, the method can analyze a reason behind taken action. We are sure that if the model returns biased results, this functionality is helpful for clearing the reason for these data bias issues.

\bibliography{emnlp2021_cameraready}
\bibliographystyle{acl_natbib}

\clearpage
\appendix
\renewcommand{\thesection}{Appendix \Alph{section}:}
\renewcommand{\thesubsection}{Appendix \Alph{section}.\arabic{subsection}:}

\section{Environment Setting}
In this section we describe Coin-Collector~\cite{Yuan+:2018:LSTM-DRQN}, a text-based game used in our experiments. Then, we describe the hyper-parameter setting. 

\subsection{Coin-Collector}
Coin-Collector is a kind of text-based games, and we have to move an agent through rooms to get a coin placed in a room.
An agent receives an observation text that describes the structure of a current room from a game. 
The goal of Coin-Collector is to analyze textual data and understand the structure of given rooms for training an agent.

A game has hyper-parameters such as {\it level} and {\it difficulty}. 
A game {\it level} indicates the minimum number of steps to a room in which a coin is placed.
Rooms are randomly connected and their structure depends on {\it difficulty}.
An easy game has no distractor rooms (dead ends) along the path. 
On a medium game, each room along the optimal trajectory has one distractor room randomly connected to it. 
A hard game, each room has two distractor rooms which means each room has one for optimal trajectory, one for the previous room, and two for distractor rooms.

An agent can use two types of verbs (\{take, go\}) and five types of nouns (\{coin, east, west, south, north\}). 
Since an action consists of a verb and a noun, there are ten different actions that an agent can take.
For the settings of LSTM-DQN++~\cite{Narasimhan+:EMNLP2015:LSTM-DQN}, the agent gets the positive reward when the agent goes in a new room.
The agent also gets positive reward when the agent successfully returns the initial coming direction for medium setting. 
If an agent takes an invalid action such as ``go coin'', or ``go north'' at no north room, the agent does not receive a negative reward.

\subsection{Hyper-parameters}
For the all experiments, we used the same hyper-parameters with the previous work for Coin-Collector as follows.
\begin{itemize}
\item We used a prioritized replay memory with capacity of $500,000$ and the priority fraction is $0.25$.  
\item A mini-batch gradient update is performed every 4 steps in the game play.  
\item The discount factor for Q-learning $\gamma$ is 0.9. 
\item We used an episodic discovery bonus that encourages an agent to discover unseen states and the coefficient $\beta$ is 1.0. 
\item We anneal the $\epsilon$ for the $\epsilon$-greedy strategy from 1 to 0.2 over 1000 epochs. After 1000 episodes,  the $\epsilon$ is 0.2.
\item We used the Adam algorithm \cite{kingma2014adam} for the optimization and the learning rate is $1e^{-3}$.
\end{itemize}

\section{Experiment details}

The training and validation times until $3,000$~epochs for each method are as follows.
\begin{itemize}
\item LSTM-DQN++~\cite{Narasimhan+:EMNLP2015:LSTM-DQN}: Around 2~hours for easy difficulty, and around 4~hours for medium difficulty.
\item NLM-DQN~\cite{dong2018neural}: Around 40~minutes for easy difficulty, and around 2.5~hours for medium difficulty.
\item NN-DQN: Around 30~minutes for easy difficulty, and around 1.5~hours for medium difficulty.
\item LNN-NN-DQN: Around 30~minutes for easy difficulty, and around 1.5~hours for medium difficulty.
\item FOL-LNN: Around 35~minutes for easy difficulty, and around 2~hours for medium difficulty.
\end{itemize}

\noindent These results are calculated on Intel~Core~i7-6700K CPU (4.00GHz) and NVIDIA Titan~X. From these results, our proposed method is not computationally expensive than other methods.

\end{document}